\documentclass{article}
\usepackage[preprint]{neurips_2026}

\usepackage[utf8]{inputenc}
\usepackage[T1]{fontenc}
\usepackage{hyperref}
\usepackage{url}
\usepackage{booktabs}
\usepackage{amsfonts}
\usepackage{amsmath}
\usepackage{amssymb}
\usepackage{nicefrac}
\usepackage{microtype}
\usepackage{xcolor}
\usepackage{multirow}
\usepackage{graphicx}
\usepackage{wrapfig}
\usepackage{makecell}

\newcommand{\pos}[1]{\textcolor{red}{#1}}
\newcommand{\negc}[1]{\textcolor{blue}{#1}}
\newcommand{\neut}[1]{#1}

\title{AIM-DDI: A Model-Agnostic Multimodal Integration Module
for Drug-Drug Interaction Prediction}

\author{
Yerin Park$^{1}$ \quad Sangseon Lee$^{1,*}$ \\
$^{1}$Department of Artificial Intelligence, Inha University \\
$^{*}$Corresponding author
}

\begin{document}

\maketitle

\begin{abstract}
Drug-drug interaction (DDI) prediction is a critical task in computational biomedicine, as adverse interactions between co-administered drugs can cause severe side effects and clinical risks.
A key challenge is unseen-drug generalization, where interactions must be predicted for drugs not observed during training.
Although multimodal DDI models exploit diverse drug-related information, their fusion mechanisms are often tied to specific prediction architectures, limiting their reuse across models.
To address this, we propose AIM-DDI, an architecture-independent multimodal integration module that represents heterogeneous modality information as tokens in a shared latent space.
By modeling dependencies across modality tokens through a unified fusion module, AIM-DDI enables model-agnostic integration of structural, chemical, and semantic drug signals across different DDI prediction architectures.
Extensive evaluations across diverse DDI models and DrugBank-based settings show that AIM-DDI consistently improves prediction performance, with the strongest gains under the most challenging both-unseen setting where neither drug in a test pair is observed during training.
These results suggest that treating multimodal integration as a reusable module, rather than a model-specific fusion component, is an effective strategy for robust unseen-drug DDI prediction.

\end{abstract}

\section{Introduction}

The co-administration of multiple drugs is common in clinical practice,
especially for patients with complex or chronic conditions.
However, incompatible drug combinations may cause adverse drug-drug interaction
(DDI) events, leading to severe side effects, reduced therapeutic efficacy, and
unexpected pharmacological risks~\citep{zitnik2018decagon,vilar2014similarity}.
Accurate DDI prediction is therefore important for safer prescription, drug
development, and pharmacovigilance.

A particularly challenging and clinically relevant setting is \emph{unseen-drug
generalization}, where a model predicts interactions involving drugs that were
not observed during training.
In this setting, prediction depends less on previously observed drug-pair
interaction patterns and more on the effective use of heterogeneous drug-related
information, including molecular structures, biological targets, enzymes,
transporters, molecular substructures, physicochemical properties, and
biomedical knowledge graphs.

This need to exploit heterogeneous information has motivated DDI models that
combine multiple sources of drug-related information.
While these models have improved predictive performance
~\citep{vilar2014similarity,ryu2018deepddi,zitnik2018decagon,deng2020ddimdl,wu2024mkgfenn},
their multimodal integration mechanisms are typically designed as part of a
specific prediction architecture.
For example, DDIMDL~\citep{deng2020ddimdl} performs feature fusion through a
dedicated multimodal network, whereas MKG-FENN~\citep{wu2024mkgfenn} integrates
multiple knowledge graph channels within an end-to-end prediction pipeline.
As a result, a fusion strategy developed for one model is difficult to apply to
another, even when similar drug modalities are available.
This limits the design of general multimodal integration methods for DDI
prediction, particularly under unseen-drug generalization.

To address this limitation, we propose AIM-DDI, a model-agnostic multimodal integration module for drug-drug interaction prediction.
AIM-DDI is designed as a multimodal integration module that is independent of a
specific prediction architecture and can be attached to different DDI models
through lightweight representation interfaces.
The key idea is to map heterogeneous modality representations into a shared
latent space as modality tokens, where dependencies across modalities can be
modeled by a unified fusion module.
This allows structural, chemical, and semantic drug information to be integrated
without redesigning the fusion mechanism for each prediction architecture.

We evaluate AIM-DDI on a DrugBank-based multimodal DDI benchmark using three
representative prediction models, DDIMDL~\citep{deng2020ddimdl},
GIL-DDI~\citep{li2026gilddi}, and MKG-FENN~\citep{wu2024mkgfenn}, with a
primary focus on one-unseen and both-unseen drug settings.
AIM-DDI yields its largest improvements when both drugs in a test pair are
unseen, with relative improvements of up to 23.34\% in accuracy, 66.04\% in macro-F1,
and 86.00\% in macro-recall across the evaluated base models.
We further conduct semantic modality ablations, evaluate AIM-DDI on additional
DrugBank-based DDI frameworks following KnowDDI~\citep{wang2024knowddi} and
KGDB-DDI~\citep{zhao2025kgdbddi}, and provide a case study on nonsteroidal
anti-inflammatory drug (NSAID)-related predictions.
Across these analyses, AIM-DDI consistently improves multimodal DDI prediction,
supporting its role as a reusable integration module for robust unseen-drug
generalization.

Our contributions are summarized as follows:
\begin{itemize}
    \item We identify architecture-dependent multimodal fusion as a key
          limitation for robust DDI prediction, especially under unseen-drug
          generalization.
    \item We propose AIM-DDI, a modality-token integration module that maps
          heterogeneous drug modalities into a shared latent space and models
          dependencies across modalities independently of a specific prediction
          architecture.
    \item We demonstrate consistent improvements across multiple DDI prediction
          models and additional DrugBank-based DDI frameworks, with particularly
          large gains when both drugs in a test pair are unseen.
\end{itemize}

\section{Related Work}

\subsection{Drug-Drug Interaction Prediction}

Computational DDI prediction has been studied through a wide range of
representations and learning paradigms.
Early approaches relied on a single modality such as drug similarity, molecular
structures, fingerprints, and pharmacological features to infer potential
interactions~\citep{vilar2014similarity,ryu2018deepddi}.
Graph-based methods later formulated DDI prediction as relational learning over
drug interaction graphs or biomedical knowledge graphs, enabling models to
capture dependencies among drugs, proteins, targets, and other biomedical
entities~\citep{zitnik2018decagon}.
More recent methods further incorporate external knowledge graphs or
drug-pair-specific subgraphs to improve prediction and interpretability
~\citep{wang2024knowddi}.
Despite these advances, many existing methods are closely tied to the
representations and architectures used to encode drug information.
This makes it difficult to separate the effect of representation learning from
the effect of multimodal integration, especially in unseen-drug settings where
models rely on heterogeneous drug-related information rather than observed
drug-pair interactions.

\subsection{Multimodal Integration for DDI Prediction}

Multimodal DDI models aim to combine complementary drug information, such as
chemical structures, biological relations, molecular substructures,
physicochemical descriptors, knowledge graphs, and textual drug descriptions.
DDIMDL integrates multiple drug features through modality-specific subnetworks
and a dedicated fusion network~\citep{deng2020ddimdl}, while MKG-FENN combines
multiple knowledge graph channels in an end-to-end DDI prediction
framework~\citep{wu2024mkgfenn}.
KGDB-DDI further fuses biological knowledge graph representations with textual
drug background information for DDI prediction~\citep{zhao2025kgdbddi}.
These studies show that heterogeneous modalities can provide complementary
signals for predicting drug interactions.
However, their fusion mechanisms are usually designed within specific model
architectures, making them difficult to reuse across different DDI prediction
models.
In contrast, our work treats multimodal integration itself as a reusable design
component and proposes AIM-DDI as an architecture-independent modality-token
integration module for robust DDI prediction under unseen-drug generalization.

\section{Method}

\begin{figure}[tb]
\centering
\includegraphics[width=0.9\linewidth]{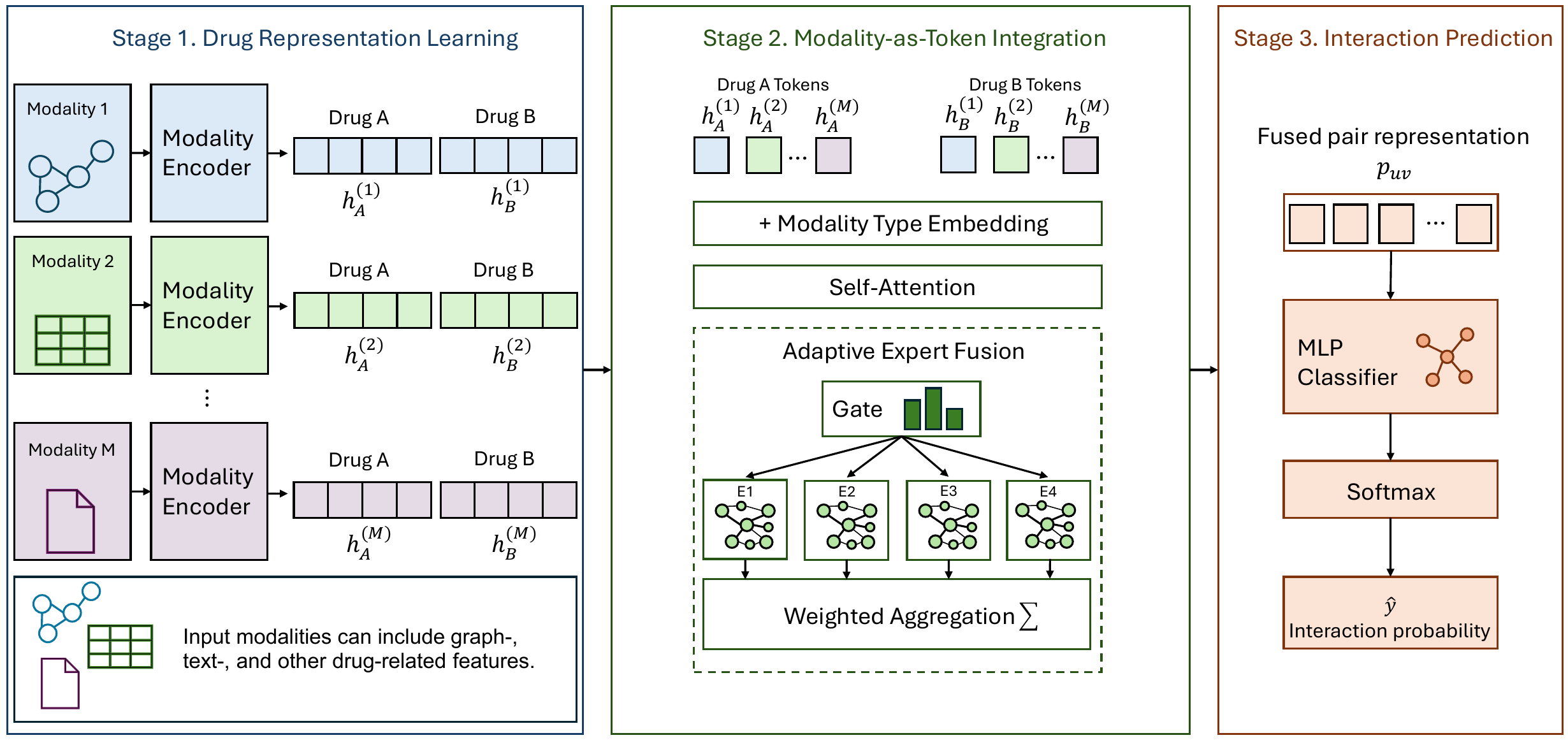}
\caption{
Overview of AIM-DDI. AIM-DDI first constructs modality-specific drug
representations from available drug-related modalities through representation
modules, then reformulates them as modality-aware tokens for drug pairs.
The token sequence is processed by modality type embedding, self-attention, and
adaptive expert fusion, and the resulting fused pair representation is used for
final DDI event prediction.
}
\label{fig:aim_overview}
\end{figure}

In this section, we present the architecture of AIM-DDI, a multimodal integration
module designed for drug-drug interaction prediction.
Figure~\ref{fig:aim_overview} illustrates the overall architecture of AIM-DDI.
AIM-DDI first constructs modality-specific drug representations from
heterogeneous drug-related data sources.
Subsequently, these representations are projected into a shared latent space and
reformulated as modality tokens.
Next, a unified fusion module models interactions among modality tokens and
produces an integrated representation of each drug pair.
Finally, the fused representation is passed to a prediction head to classify the
interaction type.

\subsection{Problem Formulation}

We consider multi-class drug-drug interaction prediction.
Let $\mathcal{D}$ denote the set of drugs and let $\mathcal{R}$ denote the set
of DDI event types, where $|\mathcal{R}|=65$ in our main benchmark.
Given a drug pair $(u,v) \in \mathcal{D} \times \mathcal{D}$, the goal is to
predict its interaction type $y_{uv} \in \mathcal{R}$.
Each drug is associated with heterogeneous drug-related information, including
structural, chemical, biological, and semantic sources.
For each available modality $m \in \mathcal{M}$, a representation module
produces modality-specific representations for the drug pair.
AIM-DDI integrates these representations and learns a prediction function $f_\theta(u,v,\{x_{uv}^{(m)}\}_{m \in \mathcal{M}}) = \hat{y}_{uv}$
, where $\hat{y}_{uv}$ is a probability distribution over DDI event types.
We evaluate the model under drug-level generalization settings, including
one-unseen and both-unseen settings, where one or both drugs in a test pair are
not observed during training.

\subsection{Drug Representation Learning}
\label{sec:drug_rep}

AIM-DDI first constructs modality-specific drug representations from
heterogeneous drug-related data sources.
The available modalities may include drug-target associations, drug-protein
relations, molecular substructures, physicochemical descriptors, drug-drug
interaction graphs, and textual drug descriptions.
Since different DDI prediction models use different subsets of these sources,
AIM-DDI does not assume a fixed modality set.
Instead, it takes modality representations produced by the underlying DDI model
or lightweight modality-specific encoders and maps them into a common hidden
space for integration.

\paragraph{Graph-based modalities.}
Many drug-related sources can be represented as graphs, where drugs are
connected to target proteins, enzymes, molecular substructures, or other drugs.
For each graph-based modality $m$, the corresponding representation module
learns trainable drug, entity, and relation embeddings.
These embeddings are randomly initialized and jointly optimized with the
prediction objective.
For drug $u$, a fixed number of neighbor-relation pairs are sampled from the
corresponding graph channel and aggregated into a drug-level representation.
Let $\mathcal{N}^{(m)}(u)$ denote the sampled neighbor-relation pairs.
We compute
\begin{equation}
    \mathbf{h}^{(m)|g}_u =
    W^{(m)}_{\mathrm{g}}
    \left[
        \mathbf{d}^{(m)}_u
        \,\Big\|\,
        \sum_{a \in \mathcal{N}^{(m)}(u)}
        \alpha^{(m)}_{u,a}\,\mathbf{e}^{(m)}_a
    \right],
\end{equation}
where $\mathbf{d}^{(m)}_u$ is the trainable drug embedding,
$\mathbf{e}^{(m)}_a$ is the trainable neighboring entity embedding,
$\alpha^{(m)}_{u,a}$ is a relation-aware attention weight, and
$W^{(m)}_{\mathrm{g}}$ projects the aggregated representation into
$\mathbb{R}^{d}$.
The initial embeddings and graph aggregation interface are inherited from the corresponding DDI model, allowing AIM-DDI to preserve model-specific structural signals while using a shared integration module.

\paragraph{LLM-based semantic modalities.}
Graph-based modalities capture explicit drug-entity relations, but their learned
embeddings may not fully reflect the biomedical semantics associated with those
relations. To complement graph representations with semantic knowledge, we
construct modality-specific textual prompts from drug-related information and
encode them using Meditron-7B, a 7B-parameter biomedical language model from the
Meditron model family~\citep{chen2023meditron,epfl2023meditron7b}.

For each drug $u$ and semantic modality $m$, multiple prompts can be generated
from the drug's associated entities, relations, substructures, or molecular
features. Each prompt is first encoded into a prompt-level embedding, and the
resulting prompt embeddings are aggregated using attention pooling to obtain one
drug-level semantic representation
$\mathbf{s}^{(m)}_u \in \mathbb{R}^{d_{\mathrm{LM}}}$ for modality $m$.

The resulting semantic embedding is projected into the shared hidden space:
\begin{equation}
    \mathbf{h}^{(m)|s}_u =
    \sigma\!\left(
        \mathrm{BN}\!\left(
            W^{(m)}_{\mathrm{s}}\mathbf{s}^{(m)}_u
        \right)
    \right)
    \in \mathbb{R}^{d},
\end{equation}
where $W^{(m)}_{\mathrm{s}}$ is a modality-specific projection matrix,
$\mathrm{BN}$ denotes batch normalization, and $\sigma$ is a nonlinear
activation function. Detailed prompt construction and attention pooling
procedures are provided in Appendix~\ref{app:llm_prompt}.

\paragraph{Pair representation construction.}
For notational simplicity, we use $\mathbf{h}^{(m)}_u$ to denote either a
graph-based representation $\mathbf{h}^{(m)|g}_u$ or a semantic representation
$\mathbf{h}^{(m)|s}_u$ when the modality type is clear from context.
Given a drug pair $(u,v)$, AIM-DDI constructs a modality-aware token sequence
from the available modality-specific representations:
\begin{equation}
    \mathbf{H}_{uv}
    =
    [
    \mathbf{h}^{(1)}_u; \cdots; \mathbf{h}^{(|\mathcal{M}_{uv}|)}_u;
    \mathbf{h}^{(1)}_v; \cdots; \mathbf{h}^{(|\mathcal{M}_{uv}|)}_v
    ]
    \in \mathbb{R}^{L \times d},
\end{equation}
where $\mathcal{M}_{uv}$ denotes the set of modalities available for the pair
and $L$ is the number of resulting tokens.
Each row of $\mathbf{H}_{uv}$ corresponds to a modality-specific representation
of one drug in the pair.
For models that already produce pair-level modality representations, AIM-DDI
uses those pair-level representations directly as tokens.
The resulting token sequence is passed to the AIM-DDI multimodal integration
module, with model-specific token construction details provided in
Appendix~\ref{app:model_interfaces}.

\subsection{Modality-as-Token Multimodal Integration}
\label{sec:modality_token}

Given the token sequence $\mathbf{H}_{uv} \in \mathbb{R}^{L \times d}$ for a
drug pair $(u,v)$, AIM-DDI integrates heterogeneous modality representations by
treating each row as a modality token.
This avoids collapsing all modalities into a single concatenated vector and
allows the model to explicitly capture dependencies among structural, chemical,
and semantic signals.

\paragraph{Modality-type embedding.}
To retain modality identity, AIM-DDI adds a learnable type embedding to each
token:
\begin{equation}
    \widetilde{\mathbf{H}}_{uv}
    =
    \mathbf{H}_{uv} + \mathbf{E}_{\mathrm{type}},
    \qquad
    \mathbf{E}_{\mathrm{type}} \in \mathbb{R}^{L \times d}.
\end{equation}
Each row of $\mathbf{E}_{\mathrm{type}}$ corresponds to the modality or
drug-channel type of the associated token.

\paragraph{Token interaction modeling.}
The type-augmented token sequence is processed by a transformer encoder block:
\begin{align}
    \mathbf{A}_{uv}
    &=
    \mathrm{LN}\!\left(
        \widetilde{\mathbf{H}}_{uv}
        +
        \mathrm{MHA}(
            \widetilde{\mathbf{H}}_{uv},
            \widetilde{\mathbf{H}}_{uv},
            \widetilde{\mathbf{H}}_{uv}
        )
    \right), \\
    \mathbf{B}_{uv}
    &=
    \mathrm{LN}\!\left(
        \mathbf{A}_{uv} + \mathrm{FFN}(\mathbf{A}_{uv})
    \right),
\end{align}
where $\mathrm{MHA}$ denotes multi-head self-attention, $\mathrm{FFN}$ is a
position-wise feed-forward network, and $\mathrm{LN}$ denotes layer
normalization.
Self-attention allows each modality token to attend to other available tokens
before fusion.
The resulting token matrix is flattened into a pair-level representation:
\begin{equation}
    \mathbf{p}_{uv} = \mathrm{Flatten}(\mathbf{B}_{uv}).
\end{equation}

\paragraph{Adaptive expert fusion.}
To allow different drug pairs to rely on different modality combinations,
AIM-DDI applies an expert-choice mixture-of-experts fusion head to
$\mathbf{p}_{uv}$.
A gating network computes expert scores:
\begin{equation}
    \mathbf{g}_{uv}
    =
    \mathrm{softmax}(W_{\mathrm{gate}}\mathbf{p}_{uv})
    \in \mathbb{R}^{E},
\end{equation}
where $E$ is the number of experts.
Under expert-choice routing, each expert selects a subset of instances according
to its routing scores under a fixed capacity.
Let $\mathcal{S}_e$ denote the set of instances selected by expert $e$.
Each expert transforms the pair representation as
\begin{equation}
    \mathbf{o}^{(e)}_{uv}
    =
    f_e(\mathbf{p}_{uv})
    =
    W^{(2)}_e
    \sigma\!\left(W^{(1)}_e \mathbf{p}_{uv} + \mathbf{b}^{(1)}_e\right)
    + \mathbf{b}^{(2)}_e .
\end{equation}
For the drug-pair instance $i$ corresponding to $(u,v)$, the final fused
representation is computed by aggregating the weighted outputs of the experts
that select the instance:
\begin{equation}
    \mathbf{z}_{i}
    =
    \frac{
        \sum_{e: i \in \mathcal{S}_e}
        g_{e,i}\, f_e(\mathbf{p}_{i})
    }{
        \sum_{e: i \in \mathcal{S}_e} g_{e,i}
    }.
\end{equation}
For notational simplicity, we denote the fused representation of the drug-pair
instance $i$ corresponding to $(u,v)$ as $\mathbf{z}_{uv}$ in the following
prediction step.
The resulting fused representation is then passed to the interaction prediction
head.

\subsection{Interaction Prediction}
\label{sec:interaction_prediction}

Given the fused representation of a drug pair $(u,v)$, AIM-DDI predicts its
interaction type with a classification head.
The classifier maps the fused representation to a probability distribution over
DDI event classes:
\begin{equation}
    \hat{\mathbf{y}}_{uv}
    =
    \mathrm{softmax}\!\left(f_{\mathrm{cls}}(\mathbf{z}_{uv})\right).
\end{equation}
The predicted label is obtained by selecting the class with the highest
probability.
In the main benchmark, the model predicts 65 DDI event classes and is optimized
using a multi-class classification objective.
Training configurations and implementation details are provided in
Appendix~\ref{app:implementation_details}.

\section{Experiments}

\subsection{Datasets}
\label{sec:main_datasets}

Following the same protocol from \cite{wu2024mkgfenn}, we evaluate AIM-DDI on a DrugBank-based multimodal DDI benchmark constructed
from DrugBank~(v5.1.7)~\citep{wishart2018drugbank} and the DDIMDL
dataset~\citep{deng2020ddimdl}.
The benchmark contains 65 DDI event classes and combines complementary
drug-related sources, including biological relations, molecular substructures,
drug-drug interaction topology, and physicochemical descriptors.
These sources provide heterogeneous structural, chemical, and semantic
information for evaluating multimodal integration.
Detailed data construction procedures and dataset statistics are provided in
Appendix~\ref{app:dataset_statistics}.

\subsection{Experimental Setup}
\label{sec:experimental_setup}

\paragraph{Task.}
We evaluate AIM-DDI under two unseen-drug generalization settings:
\emph{one-unseen}, where each test pair contains one drug that is not observed
during training, and \emph{both-unseen}, where both drugs in a test pair are
unseen during training.
We construct these settings by applying 10-fold splits at the drug level.
We additionally report the conventional seen-drug setting, based on pair-level
10-fold cross-validation, as a supplementary evaluation in
Appendix~\ref{app:seen_drug_results}.

\paragraph{Baselines.}
We evaluate AIM-DDI with three representative DDI prediction models:
DDIMDL~\citep{deng2020ddimdl}, GIL-DDI~\citep{li2026gilddi}, and
MKG-FENN~\citep{wu2024mkgfenn}.
These models cover feature-based, graph-based, and multimodal
knowledge-graph-based DDI prediction architectures.
For each model, AIM-DDI is integrated through a lightweight representation
interface while keeping the original prediction task and evaluation setting.
Model-specific integration details are provided in Appendix~\ref{app:model_interfaces}.

\paragraph{Evaluation.}
We report Accuracy, micro-AUC, micro-AUPR, macro-F1, macro-Precision, and
macro-Recall, including macro-averaged metrics to account for class imbalance.
Hyperparameters, training configurations, and compute resources are provided in
Appendix~\ref{app:implementation_details}.

\subsection{Unseen-Drug Generalization Performance}
\label{sec:unseen_drug_generalization}

Table~\ref{tab:unseen_results} reports the results under the one-unseen and
both-unseen settings, where test pairs contain drugs not observed during
training. These settings directly evaluate unseen-drug generalization. Results
for the conventional seen-drug setting are provided in
Appendix~\ref{app:seen_drug_results}, where baseline performances are already
near saturation and AIM-DDI yields relatively modest improvements.

\begin{table}[tb]
\centering
\small
\setlength{\tabcolsep}{2.5pt}
\caption{Performance comparison under unseen-drug generalization settings. Red and blue indicate increased and decreased performance relative to each baseline, respectively. Standard deviations are reported in Appendix~\ref{app:std_unseen_performance}.}
\label{tab:unseen_results}
\resizebox{\linewidth}{!}{
\begin{tabular}{lllcccccc}
\toprule
Setting & Model & Method & ACC & AUC & AUPR & F1 & PRE & REC \\
\midrule

\multirow{9}{*}{One-unseen}
& \multirow{3}{*}{DDIMDL}
& Baseline   & 0.6467 & 0.9798 & 0.6695 & 0.4756 & 0.6360 & 0.4051 \\
& & + AIM-DDI  & 0.6819 & 0.9774 & 0.7045 & 0.5375 & 0.5954 & 0.5117 \\
& & Change (\%) & \pos{+5.44\%} & \negc{-0.24\%} & \pos{+5.23\%} & \pos{+13.02\%} & \negc{-6.38\%} & \pos{+26.31\%} \\
\cmidrule(lr){2-9}

& \multirow{3}{*}{GIL-DDI}
& Baseline   & 0.7161 & 0.9558 & 0.7198 & 0.5758 & 0.6689 & 0.5269 \\
& & + AIM-DDI  & 0.7517 & 0.9753 & 0.7263 & 0.6318 & 0.7227 & 0.5856 \\
& & Change (\%) & \pos{+4.97\%} & \pos{+2.04\%} & \pos{+0.90\%} & \pos{+9.73\%} & \pos{+8.04\%} & \pos{+11.14\%} \\
\cmidrule(lr){2-9}

& \multirow{3}{*}{MKG-FENN}
& Baseline   & 0.6700 & 0.9659 & 0.6908 & 0.5313 & 0.6068 & 0.4984 \\
& & + AIM-DDI  & 0.6758 & 0.9736 & 0.6999 & 0.5443 & 0.6435 & 0.4992 \\
& & Change (\%) & \pos{+0.87\%} & \pos{+0.80\%} & \pos{+1.32\%} & \pos{+2.45\%} & \pos{+6.05\%} & \pos{+0.16\%} \\
\midrule

\multirow{9}{*}{Both-unseen}
& \multirow{3}{*}{DDIMDL}
& Baseline   & 0.4262 & 0.9503 & 0.3824 & 0.1381 & 0.2316 & 0.1193 \\
& & + AIM-DDI  & 0.4788 & 0.9503 & 0.4417 & 0.2293 & 0.2724 & 0.2219 \\
& & Change (\%) & \pos{+12.34\%} & \neut{+0.00\%} & \pos{+15.51\%} & \pos{+66.04\%} & \pos{+17.62\%} & \pos{+86.00\%} \\
\cmidrule(lr){2-9}

& \multirow{3}{*}{GIL-DDI}
& Baseline   & 0.5449 & 0.9081 & 0.4836 & 0.2806 & 0.3897 & 0.2492 \\
& & + AIM-DDI  & 0.5812 & 0.9389 & 0.4801 & 0.3353 & 0.4643 & 0.3209 \\
& & Change (\%) & \pos{+6.66\%} & \pos{+3.39\%} & \negc{-0.72\%} & \pos{+19.49\%} & \pos{+19.14\%} & \pos{+28.77\%} \\
\cmidrule(lr){2-9}

& \multirow{3}{*}{MKG-FENN}
& Baseline   & 0.4623 & 0.9136 & 0.4253 & 0.2175 & 0.2735 & 0.2122 \\
& & + AIM-DDI  & 0.5702 & 0.9606 & 0.5085 & 0.3267 & 0.4032 & 0.3015 \\
& & Change (\%) & \pos{+23.34\%} & \pos{+5.14\%} & \pos{+19.56\%} & \pos{+50.21\%} & \pos{+47.42\%} & \pos{+42.08\%} \\

\bottomrule
\end{tabular}
}
\end{table}

AIM-DDI improves most metrics across DDIMDL, GIL-DDI, and MKG-FENN, indicating
that the proposed integration module is effective across feature-based,
graph-based, and multimodal knowledge-graph-based prediction models. The gains
are stronger in the both-unseen setting, where generalization is most
challenging because neither drug in the test pair is observed during training.
Across the evaluated base models, AIM-DDI achieves relative improvements of up
to 23.34\% in ACC, 19.56\% in AUPR, 66.04\% in macro-F1, and 86.00\% in
macro-Recall.

The improvements are not tied to a single prediction architecture. In the
one-unseen setting, AIM-DDI improves most metrics for all three base models. In
the both-unseen setting, DDIMDL benefits substantially in macro-F1 and
macro-Recall, while MKG-FENN improves across all metrics, including 23.34\% in
ACC and 50.21\% in macro-F1. Although some ranking-based metrics show small
decreases, such as AUPR for GIL-DDI in the both-unseen setting, the consistent
gains in macro-averaged metrics suggest that AIM-DDI improves class-balanced
prediction quality under challenging unseen-drug conditions.

\subsection{Ablation Studies}
\label{sec:ablation}

We conduct ablation studies in the both-unseen setting using the MKG-FENN-based
AIM-DDI implementation. All variants use the same training and inference
configuration unless otherwise specified. The goal is to examine the effects of (1) semantic modality selection, 
(2) semantic encoder choice, (3) expert routing, 
(4) the number of experts, and (5) pair representation construction.
Due to the limited space, experimental results of (4)-(5) are reported in Appendix~\ref{app:expert_number} and \ref{app:pair_representation}, respectively.

\begin{table}[tb]
\centering
\small
\setlength{\tabcolsep}{3.2pt}
\caption{
Ablation study of LLM-derived semantic modalities in the both-unseen setting.
A check mark indicates that the corresponding semantic modality is used.
Replacement variants substitute selected graph channels with LLM-derived representations, while Parallel GNN+LLM uses graph and LLM representations jointly. Standard deviations are reported in Appendix~\ref{app:std_ablation_modality}.}
\label{tab:llm_ablation}
\resizebox{\linewidth}{!}{
\begin{tabular}{lccc|ccccccc}
\toprule
Variant & BioRel & MolSub & DDIGraph & ACC & AUC & AUPR & F1 & Pre & Rec & F-rank \\
\midrule
None &  &  &  & 0.5046 & 0.9492 & 0.4560 & 0.2397 & 0.3001 & 0.2272 & 6.50 \\
BioRel & $\checkmark$ &  &  & 0.4613 & 0.9411 & 0.4096 & 0.2185 & 0.2557 & 0.2198 & 4.67 \\
MolSub &  & $\checkmark$ &  & 0.5309 & 0.9560 & 0.4917 & 0.2740 & 0.3666 & 0.2472 & 8.00 \\
DDIGraph &  &  & $\checkmark$ & 0.3637 & 0.9228 & 0.2927 & 0.1516 & 0.1719 & 0.1566 & 2.00 \\
BioRel + DDIGraph & $\checkmark$ &  & $\checkmark$ & 0.3263 & 0.9127 & 0.2607 & 0.1536 & 0.1911 & 0.1534 & 1.33 \\
MolSub + DDIGraph &  & $\checkmark$ & $\checkmark$ & 0.3640 & 0.9219 & 0.2841 & 0.1680 & 0.2086 & 0.1638 & 2.67 \\
All semantic & $\checkmark$ & $\checkmark$ & $\checkmark$ & 0.4591 & 0.9472 & 0.4100 & 0.1972 & 0.2512 & 0.1873 & 4.33 \\
Parallel GNN+LLM & $\checkmark$ & $\checkmark$ &  & 0.4688 & 0.9474 & 0.4277 & 0.2468 & 0.3140 & 0.2369 & 6.50 \\
\textbf{BioRel + MolSub$^{*}$} & $\checkmark$ & $\checkmark$ &  & \textbf{0.5702} & \textbf{0.9606} & \textbf{0.5085} & \textbf{0.3267} & \textbf{0.4032} & \textbf{0.3015} & \textbf{9.00} \\
\bottomrule
\end{tabular}
}
\vspace{1mm}
\footnotesize{
$^{*}$BioRel + MolSub is used as the proposed AIM-DDI configuration.
}
\end{table}

\subsubsection{Effect of LLM-Derived Semantic Modalities}
\label{sec:ablation_LLM_modal}

We analyze whether LLM-derived semantic modalities provide complementary
signals to graph-based representations. BioRel, MolSub, and DDIGraph denote
semantic representations constructed from biological relation, molecular
substructure, and DDI graph contexts, respectively. In replacement variants,
selected semantic sources replace the corresponding graph channels, while
unselected channels remain graph-based. Parallel GNN+LLM uses both
representations simultaneously. Detailed experimental setups are provided in
Appendix~\ref{app:llm_ablation_details}.

As shown in Table~\ref{tab:llm_ablation}, BioRel + MolSub achieves the best
performance across all metrics, improving ACC, AUPR, and macro-F1 over the
no-LLM setting. Parallel GNN+LLM does not outperform this configuration,
suggesting that selective semantic integration is more effective than simply
combining graph and LLM representations. The poor performance of
DDIGraph-related variants further indicates that not all semantic sources are
equally aligned with the DDI prediction objective.

\begin{wraptable}{r}{0.43\linewidth}
\vspace{-25pt}
\centering
\small
\caption{Ablation study results for the effect of LLM-based semantic encoder choice. Standard deviations are reported in Appendix~\ref{app:std_ablation_LLM}.}
\label{tab:semantic_encoder}
\vspace{2mm}
\begin{tabular}{lccc}
\toprule
Encoder & ACC & AUC & F1 \\
\midrule
Meditron-7B & 0.5702 & 0.9606 & 0.3267 \\
BioBERT     & 0.5686 & 0.9582 & 0.3296 \\
\bottomrule
\end{tabular}
\end{wraptable}

\subsubsection{Semantic Encoder Analysis}
\label{sec:semantic_encoder_analysis}

We compare Meditron-7B~\citep{chen2023meditron,epfl2023meditron7b} and
BioBERT~\citep{lee2020biobert} as semantic encoders in the both-unseen setting.
As shown in Table~\ref{tab:semantic_encoder}, the two encoders yield comparable
performance, with Meditron slightly better in ACC/AUC and BioBERT slightly
better in F1. This suggests that the gains mainly come from multimodal
integration rather than a specific LLM encoder.

\subsubsection{Expert Routing and Modality Contribution}
\label{sec:expert_routing}

We analyze whether the expert-choice fusion module routes drug-pair
representations according to different event and modality patterns in the
both-unseen setting. As shown in Table~\ref{tab:expert_summary}, samples are
distributed across all four experts, indicating that routing does not collapse
to a single expert.

\begin{table}[tb]
\centering
\small
\caption{
Ablation study results for expert routing patterns in the both-unseen setting.
Top events denote the most frequent DDI event classes assigned to each expert.
Semantic ratio is computed as BioRel + MolSub. 
}
\label{tab:expert_summary}
\resizebox{\linewidth}{!}{
\begin{tabular}{lcccccc}
\toprule
Expert & Samples & Top-1 event & Top-2 event & BioRel & MolSub & Semantic ratio \\
\midrule
Expert 0
& 802
& Class 0 (metabolism decrease; 36.66\%)
& Class 1 (adverse effects increase; 21.32\%)
& 0.1425 & 0.0555 & 0.1980 \\
Expert 1
& 1,059
& Class 0 (metabolism decrease; 28.99\%)
& Class 1 (adverse effects increase; 17.85\%)
& 0.1276 & 0.0699 & 0.1975 \\
Expert 2
& 1,107
& Class 1 (adverse effects increase; 23.40\%)
& Class 0 (metabolism decrease; 23.22\%)
& 0.1199 & 0.0730 & 0.1929 \\
Expert 3
& 752
& Class 1 (adverse effects increase; 40.82\%)
& Class 0 (metabolism decrease; 16.49\%)
& 0.1126 & 0.0822 & 0.1948 \\
\bottomrule
\end{tabular}
}
\end{table}

The routing results show event-level differences across experts. Experts 0 and
1 are more frequently associated with metabolism decrease events, whereas
Experts 2 and 3 are more associated with adverse-effect increase events. Expert
3 shows the strongest concentration on adverse-effect increase, while Expert 2
has a more balanced distribution between the two dominant event types.

The modality contribution results show that DDIGraph remains the dominant
structural signal across experts. However, the relative use of semantic
modalities differs: BioRel is higher in Experts 0 and 1, while MolSub becomes
relatively more prominent in Experts 2 and 3. These observations suggest that
the routing module assigns drug pairs to experts with different event tendencies
and semantic-modality profiles. Full routing statistics are provided in
Appendix~\ref{app:expert_routing_details}.

\subsection{Adaptation to Additional DrugBank-Based DDI Frameworks}
\label{sec:additional_drugbank_frameworks}

We further evaluate AIM-DDI under two additional DrugBank-based DDI frameworks
following KnowDDI~\citep{wang2024knowddi} and
KGDB-DDI~\citep{zhao2025kgdbddi}.
Although these settings are also derived from DrugBank, they differ from the
main benchmark in data construction and evaluation protocols. Detailed
configurations are provided in Appendix~\ref{app:additional_frameworks}.

\begin{table}[tb]
\centering
\small
\setlength{\tabcolsep}{3pt}
\caption{
Adaptation results on additional DrugBank-based DDI frameworks.
AIM-DDI is applied to KnowDDI and KGDB-DDI to evaluate
whether the proposed integration module provides gains across different DrugBank-derived model configurations.
Change (\%) denotes the relative improvement over the corresponding baseline.}
\label{tab:additional_drugbank_frameworks}
\begin{tabular}{llcccccc}
\toprule
Framework & Method & ACC & AUC & AUPR & F1 & PRE & REC \\
\midrule

\multirow{3}{*}{KnowDDI-based}
& Baseline   & 0.9183 & 0.9988 & 0.9350 & 0.8805 & 0.8984 & 0.8739 \\
& + AIM-DDI  & 0.9770 & 0.9998 & 0.9716 & 0.8927 & 0.9048 & 0.8936 \\
\cmidrule(lr){2-8}
& Change (\%)  & \pos{+6.39\%} & \pos{+0.10\%} & \pos{+3.91\%} & \pos{+1.39\%} & \pos{+0.71\%} & \pos{+2.25\%} \\

\midrule

\multirow{3}{*}{KGDB-DDI-based}
& Baseline   & 0.9749 & 0.9952 & 0.9955 & 0.9749 & 0.9766 & 0.9731 \\
& + AIM-DDI  & 0.9780 & 0.9974 & 0.9974 & 0.9779 & 0.9805 & 0.9753 \\
\cmidrule(lr){2-8}
& Change (\%)  & \pos{+0.32\%} & \pos{+0.22\%} & \pos{+0.19\%} & \pos{+0.31\%} & \pos{+0.40\%} & \pos{+0.23\%} \\

\bottomrule
\end{tabular}
\end{table}

As shown in Table~\ref{tab:additional_drugbank_frameworks}, AIM-DDI improves
all metrics in both settings. The gains are larger in the KnowDDI-based
framework, while the KGDB-DDI-based framework shows smaller but consistent
improvements due to its already strong baseline. These results suggest that
AIM-DDI provides additional benefit across different DrugBank-based settings
without relying on a specific model configuration.

\subsection{Case Study: Analysis of NSAID-related Predictions}
\label{sec:nsaid_case_study}

We conduct a case study on nonsteroidal anti-inflammatory drug
(NSAID)-related pairs in the both-unseen setting. NSAIDs are widely used for
pain and inflammation management, but are also associated with gastrointestinal,
cardiovascular, renal, and drug-drug interaction risks
~\citep{wongrakpanich2018nsaid,moore2015nsaid}. Because NSAIDs are often
co-prescribed with potentially interacting drugs, they form a clinically
meaningful subset for DDI prediction~\citep{abdu2020nsaid}.

\begin{figure}[tb]
\centering
\includegraphics[width=0.85\linewidth]{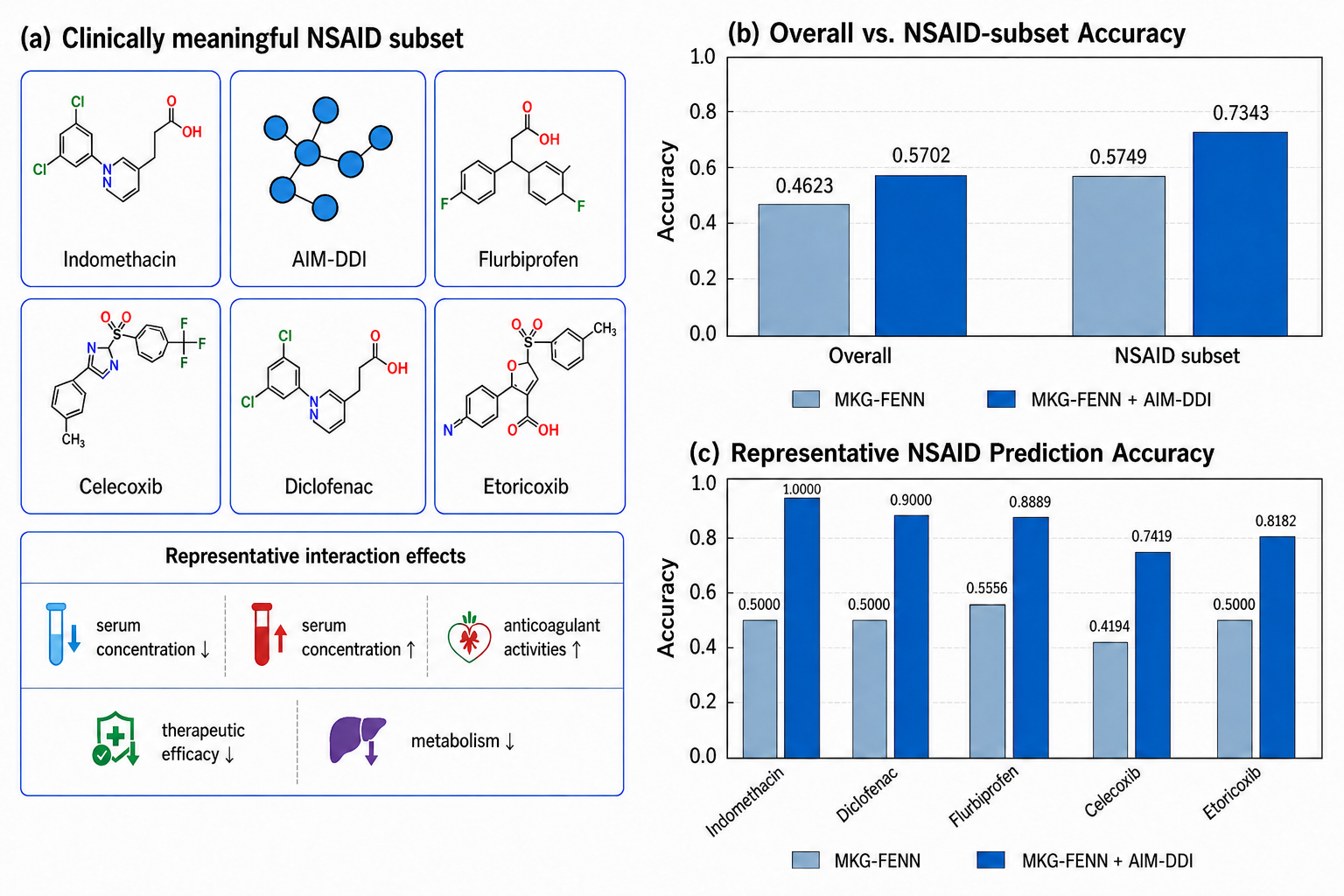}
\caption{
Case study on NSAID-related predictions in the both-unseen setting.
(a) Qualitative illustration of the clinically meaningful NSAID subset and
representative interaction-effect categories.
(b) Overall and NSAID-subset accuracy comparison between MKG-FENN and
MKG-FENN + AIM-DDI.
(c) Prediction accuracy for five representative NSAIDs.
}
\label{fig:nsaid_case_study}
\end{figure}

Figure~\ref{fig:nsaid_case_study}(a) illustrates representative NSAID-related
interaction patterns, including serum concentration changes, metabolism
decrease, anticoagulant activity increase, and therapeutic efficacy decrease.
This shows that the NSAID subset involves diverse interaction patterns rather
than a single homogeneous interaction type.

As shown in Figure~\ref{fig:nsaid_case_study}(b), AIM-DDI improves overall
both-unseen accuracy from 0.4623 to 0.5702 and NSAID-subset accuracy from
0.5749 to 0.7343, corresponding to absolute differences of 0.1064 and 0.1594,
respectively. Figure~\ref{fig:nsaid_case_study}(c) further shows large
improvements for several representative NSAIDs, including Indomethacin,
Diclofenac, Flurbiprofen, Celecoxib, and Etoricoxib.

Although the gains are not uniform across all NSAIDs, the overall pattern shows
that AIM-DDI improves prediction behavior on a clinically relevant subset of
both-unseen drug pairs.

\section{Conclusion}

In this work, we proposed AIM-DDI, a model-agnostic multimodal integration
module for drug-drug interaction prediction.
AIM-DDI reformulates heterogeneous modality representations as tokenized
representations and integrates them through a unified fusion framework that can
be connected to different DDI prediction models through lightweight
representation interfaces.
Experimental results across multiple DDI prediction models show that the
proposed framework is particularly effective in unseen-drug generalization
settings, where one or both drugs in a test pair are not observed during
training.
These findings suggest that multimodal integration should be treated as an
independent and reusable design component for robust DDI prediction.

In future work, we plan to investigate more expressive modality interaction
mechanisms, explore dynamic modality selection for settings where only a subset
of modalities is available at inference time, and extend AIM-DDI to broader
biomedical interaction prediction tasks such as drug-target interaction and
protein-protein interaction prediction.

\bibliographystyle{plainnat}
\bibliography{references}

\newpage
\clearpage
\appendix
\section*{Appendix}

\setcounter{table}{0}
\renewcommand{\thetable}{A\arabic{table}}
\renewcommand{\theHtable}{appendix.table.A\arabic{table}}

\setcounter{figure}{0}
\renewcommand{\thefigure}{A\arabic{figure}}
\renewcommand{\theHfigure}{appendix.figure.A\arabic{figure}}

\section{Dataset Construction and Statistics}
\label{app:dataset_statistics}

Following the same protocol from \cite{wu2024mkgfenn}, the main benchmark is constructed from DrugBank~(v5.1.7)~\citep{wishart2018drugbank} and the DDIMDL dataset~\citep{deng2020ddimdl}.
It consists of four complementary components. Each component is stored
in the form of \textit{drug // entity // relation} or
\textit{drug // feature // value}, and is converted into a knowledge-graph
triplet following the convention \textit{(head, relation, tail)}.
The set of available modalities may differ across experimental settings.

\paragraph{Drug-protein knowledge graph.}
The first component is derived from DrugBank and captures biological
relationships between drugs and protein-related entities, such as targets,
transporters, binding proteins, and enzymes. For each drug-entity association,
the general biological function of the entity is used as the relation type.
Thus, each record is represented as a triplet of the form
\textit{(drug, function, protein/entity)}. For example, the file entry
\textit{Lovastatin // Serum albumin // Toxic substance binding} is represented
as \textit{(Lovastatin, Toxic substance binding, Serum albumin)}. This component
captures indirect biological mechanisms that may be relevant to drug-drug
interactions.

\paragraph{Molecular substructure knowledge graph.}
The second component is constructed from SMILES-derived molecular substructures
in the DDIMDL dataset. Each drug is decomposed into a set of molecular
substructure identifiers, and the relation between a drug and each substructure
is represented as \textit{include}. Each triplet is represented as
\textit{(drug, include, substructure)}. For example, an entry such as
\textit{Glucosamine // 9 // include} indicates that Glucosamine includes the
molecular substructure indexed by 9. This component provides fine-grained
chemical substructure information for each drug.

\paragraph{Drug-drug interaction knowledge graph.}
The third component is derived from the DDI event matrix. Each known drug-drug
interaction is converted into a graph triplet where the two drugs correspond to
the head and tail entities, and the DDI event type corresponds to the relation.
Each triplet is represented as \textit{(drug\_A, interaction, drug\_B)}. This
component provides direct interaction topology and serves as the primary
structural signal for DDI event prediction.

\paragraph{Molecular descriptor features.}
The fourth component contains molecular descriptor and fingerprint information,
including physicochemical properties such as molecular weight, MolLogP, and
selected MACCS fingerprints. Each file entry is stored as
\textit{drug // feature // value}. Following the triplet convention, this can be
represented as \textit{(drug, value, feature)}. For example, the entry
\textit{Glucosamine // NumSaturatedRings // 3} is represented as
\textit{(Glucosamine, 3, NumSaturatedRings)}. This component provides global
chemical properties and descriptor-level information that are not explicitly
captured by graph topology.

In the main unseen-drug experiments, graph-structured modalities are primarily
used for token construction, while descriptor information is retained as part of
the benchmark description and used when supported by the corresponding base DDI
prediction model. This setting allows AIM-DDI to focus on consistently available
drug-level structural and semantic representations across the evaluated DDI
prediction models.

Detailed statistics for the graph-structured components are reported in
Table~\ref{tab:dataset_statistics}.

\begin{table}[b]
\centering
\small
\caption{Statistics of the multimodal DDI datasets.}
\label{tab:dataset_statistics}
\begin{tabular}{lcccc}
\toprule
Component & Drug & Entity & Relation & Triples \\
\midrule
Drug-protein KG & 572 & 825 & 235 & 6,541 \\
Molecular substructure KG & 572 & 583 & 1 & 70,350 \\
Drug-drug interaction KG & 561 & 558 & 65 & 37,264 \\
Molecular descriptors & 572 & 20 & scalar value & 11,440 \\
\bottomrule
\end{tabular}
\end{table}

\section{LLM-based Semantic Representation Construction}
\label{app:llm_prompt}

To construct LLM-based semantic representations, we convert drug-level and
modality-specific information into textual prompts and encode them using
Meditron-7B\citep{epfl2023meditron7b}. The purpose of this process is to obtain a semantic embedding for
each drug and each modality, analogous to how graph-based encoders aggregate
multiple connected entities into a single drug-level representation.

\paragraph{Drug identity prompt.}
For each drug, we first construct a basic drug identity prompt of the form:
\begin{quote}
\small
\texttt{A drug is [DRUG].}
\end{quote}
For example, for the drug Herceptin, the prompt is:
\begin{quote}
\small
\texttt{A drug is Herceptin.}
\end{quote}
This prompt provides a simple semantic representation of the drug name itself.

\paragraph{Modality-specific prompts.}
For each modality, we construct textual prompts from the corresponding
drug-related information. If a drug is connected to multiple entities in a
modality, each relation-entity pair is converted into a sentence-level prompt.

For biological relation information, we use prompts of the form:
\begin{quote}
\small
\texttt{A drug is associated with target gene [GENE].}
\end{quote}
The same template can be used for protein-related entities by replacing
\texttt{[GENE]} with the corresponding protein, transporter, enzyme, or
biological entity.

For molecular descriptor information, we use prompts such as:
\begin{quote}
\small
\texttt{A drug has molecular weight [VALUE].}
\end{quote}
For example, if the molecular weight value is discretized as 4, the prompt is:
\begin{quote}
\small
\texttt{A drug has molecular weight 4.}
\end{quote}

For molecular substructure information, a drug can be connected to multiple
substructure identifiers. In this case, each substructure connection is converted
into a separate prompt:
\begin{quote}
\small
\texttt{A drug includes molecular substructure [SUBSTRUCTURE].}
\end{quote}
Thus, if a drug is connected to several substructures, multiple prompts are
generated for that drug under the molecular substructure modality.

For drug-drug interaction graph contexts, neighboring drugs and their
interaction relation types are also converted into textual prompts:
\begin{quote}
\small
\texttt{A drug has interaction [RELATION] with drug [DRUG].}
\end{quote}

\paragraph{From multiple prompts to one drug-level modality embedding.}
A single drug can be associated with multiple entities within the same modality.
For instance, one drug may be connected to multiple molecular substructures or
multiple target genes. To obtain one semantic representation per drug and per
modality, we first encode each prompt independently with Meditron-7B. Let
$\mathcal{T}^{(m)}_u = \{t^{(m)}_{u,1}, \ldots, t^{(m)}_{u,K}\}$ denote the set
of textual prompts generated for drug $u$ under modality $m$. Each prompt
$t^{(m)}_{u,k}$ is encoded by Meditron-7B and converted into a fixed-size prompt
embedding $\mathbf{e}^{(m)}_{u,k}$.

Since not all prompts associated with a drug are equally informative, we apply
attention pooling over the prompt embeddings within the same modality. For each
prompt embedding $\mathbf{e}^{(m)}_{u,k}$, an attention score is computed as
\begin{equation}
    a^{(m)}_{u,k}
    =
    \mathbf{q}^{(m)\top}
    \tanh\!\left(
        W^{(m)}_{\mathrm{att}}\mathbf{e}^{(m)}_{u,k}
    \right),
\end{equation}
where $W^{(m)}_{\mathrm{att}}$ and $\mathbf{q}^{(m)}$ are learnable attention
parameters for modality $m$. The attention weights are then normalized across
all prompts of the same drug and modality:
\begin{equation}
    \alpha^{(m)}_{u,k}
    =
    \frac{\exp(a^{(m)}_{u,k})}
    {\sum_{j=1}^{K}\exp(a^{(m)}_{u,j})}.
\end{equation}
The final LLM-based semantic representation of drug $u$ for modality $m$ is
obtained as the weighted sum of prompt embeddings:
\begin{equation}
    \mathbf{s}^{(m)}_u
    =
    \sum_{k=1}^{K}
    \alpha^{(m)}_{u,k}
    \mathbf{e}^{(m)}_{u,k}.
\end{equation}
This produces one modality-specific semantic embedding for each drug, while
allowing the model to emphasize more informative prompts among multiple
drug-related entities or substructures.

\paragraph{Projection to the AIM-DDI hidden space.}
The resulting modality-specific semantic embedding $\mathbf{s}^{(m)}_u$ is
projected into the same hidden dimension as the graph-based representation:
\begin{equation}
    \mathbf{h}^{(m)|s}_u =
    \sigma\!\left(
        \mathrm{BN}\!\left(
            W^{(m)}_{\mathrm{s}}\mathbf{s}^{(m)}_u
        \right)
    \right),
\end{equation}
where $W^{(m)}_{\mathrm{s}}$ is a modality-specific projection matrix,
$\mathrm{BN}$ denotes batch normalization, and $\sigma$ denotes a nonlinear
activation function. The projected representation is then used as a semantic
modality token in AIM-DDI.

\section{Integration Details for Base DDI Prediction Models}
\label{app:model_interfaces}

AIM-DDI is applied to each base DDI prediction model through a lightweight
representation interface. The goal of this interface is to convert the available
model-specific representations into modality-aware tokens while preserving the
original input representation and prediction setting of each model.

\paragraph{DDIMDL.}
DDIMDL~\citep{deng2020ddimdl} is a feature-based multimodal framework that constructs multiple drug
features from biological and chemical attributes and combines them through
modality-specific subnetworks and a joint fusion layer. In our implementation,
the pair-level feature representations produced from each modality are projected
into a shared hidden dimension and used as modality tokens for AIM-DDI. Since
DDIMDL operates on pair-level handcrafted features, LLM-derived semantic
embeddings are used as an auxiliary semantic guidance signal for the fusion
module rather than as separate drug-level graph tokens.

\paragraph{GIL-DDI.}
GIL-DDI~\citep{li2026gilddi} learns drug representations from multiple graph views using
attention-based neighbor aggregation. In our implementation, the graph-view
representations of the two drugs in a pair are stacked as drug-level channel
tokens. With three graph channels, a pair $(u,v)$ is represented as
\[
    \{\mathbf{h}^{1}_{u}, \mathbf{h}^{2}_{u}, \mathbf{h}^{3}_{u},
    \mathbf{h}^{1}_{v}, \mathbf{h}^{2}_{v}, \mathbf{h}^{3}_{v}\}.
\]
The resulting sequence is augmented with modality-type embeddings and passed to
the AIM-DDI fusion module.

\paragraph{MKG-FENN.}
MKG-FENN~\citep{wu2024mkgfenn} encodes drugs through multiple knowledge graph channels and fuses the
resulting representations for DDI prediction. In our implementation, the
per-drug channel representations are converted into modality-aware tokens in the
same drug-level sequence format as GIL-DDI. AIM-DDI then models interactions
among the channel tokens and produces the final prediction.

\section{Details of LLM-derived Semantic Modality Ablation}
\label{app:llm_ablation_details}

We provide additional details for the LLM-derived semantic modality ablation in
Section~\ref{sec:ablation_LLM_modal}. This ablation is conducted in the
both-unseen setting using the MKG-FENN-based AIM-DDI implementation. All
variants use the same data split, training objective, optimizer, number of
experts, routing strategy, and evaluation protocol. Only the semantic modality
configuration is changed.

\paragraph{Semantic modality definitions.}
BioRel denotes the LLM-derived semantic representation constructed from
drug-protein biological relation contexts. MolSub denotes the semantic
representation constructed from molecular substructure contexts. DDIGraph
denotes the semantic representation constructed from drug-drug interaction graph
contexts. Each semantic representation is generated from modality-specific
textual prompts and projected into the same hidden dimension as the
corresponding graph-based representation.

\paragraph{Replacement setting.}
In the replacement variants, selected semantic modalities replace their
corresponding graph-based representation channels. For example, the BioRel
variant replaces the drug-protein graph representation with the BioRel semantic
representation, while the remaining channels are kept graph-based. Similarly,
BioRel + MolSub replaces the drug-protein and molecular substructure channels
with their semantic counterparts, while retaining the DDIGraph structural
channel.

\paragraph{Parallel GNN+LLM setting.}
The Parallel GNN+LLM variant uses both graph-based and LLM-derived
representations simultaneously rather than replacing graph channels. This
setting is included to examine whether simply increasing the number of
representations improves performance. The results show that parallel use of GNN
and LLM representations does not outperform the selective replacement strategy,
suggesting that indiscriminately adding semantic sources may introduce redundant
or weakly aligned information.

\paragraph{Interpretation.}
The ablation results indicate that the effectiveness of LLM-derived semantic
modalities depends on the source modality. BioRel and MolSub provide
complementary signals to graph-based representations, whereas DDIGraph-related
semantic variants show weaker performance. This suggests that semantic
representations are most useful when they complement structural information
rather than duplicate or blur the original DDI graph signal.

\section{Standard Deviation of Unseen-Drug Generalization Performance}
\label{app:std_unseen_performance}

We further analyze fold-level variability of the unseen-drug generalization
performance. Table~\ref{tab:std_unseen_performance} reports the mean and
standard deviation over 10-fold cross-validation, where the standard deviation
captures variability across different drug-level train/test splits.

Across most settings, AIM-DDI consistently improves the mean performance
compared with the corresponding baselines. The standard deviation is reduced
in some settings, while slightly increased in others, indicating that the
performance gain is not solely associated with reduced variance across folds.
Importantly, even when the standard deviation increases, the mean improvement
remains consistent, suggesting that the observed gains are not driven by a
single favorable split.

The results also show that the both-unseen setting generally exhibits larger
variance than the one-unseen setting, reflecting the increased difficulty of
generalizing to completely unseen drug pairs.

\begin{table}[b]
\centering
\small
\caption{
Mean and standard deviation of unseen-drug generalization performance.
Results are reported as mean $\pm$ standard deviation when fold-level values are
available.
}
\label{tab:std_unseen_performance}
\resizebox{\linewidth}{!}{
\begin{tabular}{lllcccccc}
\toprule
Setting & Model & Method & ACC & AUC & AUPR & F1 & PRE & REC \\
\midrule

\multirow{6}{*}{One-unseen}
& \multirow{2}{*}{DDIMDL}
& Baseline
& 0.6467 $\pm$ 0.0352
& 0.9798 $\pm$ 0.0029
& 0.6695 $\pm$ 0.0347
& 0.4756 $\pm$ 0.0395
& 0.6360 $\pm$ 0.0372
& 0.4051 $\pm$ 0.0369 \\
& & + AIM-DDI
& 0.6819 $\pm$ 0.0495
& 0.9774 $\pm$ 0.0222
& 0.7045 $\pm$ 0.0632
& 0.5375 $\pm$ 0.0704
& 0.5954 $\pm$ 0.0765
& 0.5117 $\pm$ 0.0743 \\
\cmidrule(lr){2-9}

& \multirow{2}{*}{GIL-DDI}
& Baseline
& 0.7161 $\pm$ 0.0345
& 0.9558 $\pm$ 0.0148
& 0.7198 $\pm$ 0.0470
& 0.5758 $\pm$ 0.0342
& 0.6689 $\pm$ 0.0460
& 0.5269 $\pm$ 0.0415 \\
& & + AIM-DDI
& 0.7517 $\pm$ 0.0461
& 0.9753 $\pm$ 0.0083
& 0.7263 $\pm$ 0.0562
& 0.6318 $\pm$ 0.0301
& 0.7227 $\pm$ 0.0479
& 0.5856 $\pm$ 0.0382 \\
\cmidrule(lr){2-9}

& \multirow{2}{*}{MKG-FENN}
& Baseline
& 0.6700 $\pm$ 0.0411
& 0.9659 $\pm$ 0.0085
& 0.6908 $\pm$ 0.0566
& 0.5313 $\pm$ 0.0492
& 0.6068 $\pm$ 0.0634
& 0.4984 $\pm$ 0.0384 \\
& & + AIM-DDI
& 0.6758 $\pm$ 0.0499
& 0.9736 $\pm$ 0.0070
& 0.6999 $\pm$ 0.0632
& 0.5443 $\pm$ 0.0475
& 0.6435 $\pm$ 0.0507
& 0.4992 $\pm$ 0.0507 \\

\midrule

\multirow{6}{*}{Both-unseen}
& \multirow{2}{*}{DDIMDL}
& Baseline
& 0.4262 $\pm$ 0.0583
& 0.9503 $\pm$ 0.0080
& 0.3824 $\pm$ 0.0563
& 0.1381 $\pm$ 0.0179
& 0.2316 $\pm$ 0.0327
& 0.1193 $\pm$ 0.0159 \\
& & + AIM-DDI
& 0.4788 $\pm$ 0.0583
& 0.9503 $\pm$ 0.0103
& 0.4417 $\pm$ 0.0662
& 0.2293 $\pm$ 0.0623
& 0.2724 $\pm$ 0.0709
& 0.2219 $\pm$ 0.0636 \\
\cmidrule(lr){2-9}

& \multirow{2}{*}{GIL-DDI}
& Baseline
& 0.5449 $\pm$ 0.0705
& 0.9081 $\pm$ 0.0236
& 0.4836 $\pm$ 0.0925
& 0.2806 $\pm$ 0.0750
& 0.3897 $\pm$ 0.0775
& 0.2492 $\pm$ 0.0763 \\
& & + AIM-DDI
& 0.5812 $\pm$ 0.0790
& 0.9389 $\pm$ 0.0092
& 0.4801 $\pm$ 0.1096
& 0.3353 $\pm$ 0.0650
& 0.4643 $\pm$ 0.0683
& 0.3209 $\pm$ 0.0659 \\
\cmidrule(lr){2-9}

& \multirow{2}{*}{MKG-FENN}
& Baseline
& 0.4623 $\pm$ 0.0533
& 0.9136 $\pm$ 0.0206
& 0.4253 $\pm$ 0.0609
& 0.2175 $\pm$ 0.0320
& 0.2735 $\pm$ 0.0376
& 0.2122 $\pm$ 0.0266 \\
& & + AIM-DDI
& 0.5702 $\pm$ 0.0814
& 0.9606 $\pm$ 0.0077
& 0.5085 $\pm$ 0.0962
& 0.3267 $\pm$ 0.0693
& 0.4032 $\pm$ 0.0724
& 0.3015 $\pm$ 0.0729 \\

\bottomrule
\end{tabular}
}
\end{table}

\section{Results on the Seen-Drug Setting}
\label{app:seen_drug_results}

Table~\ref{tab:seen_drug_results} reports the results on the seen-drug setting,
where both drugs in each test pair can appear during training. Compared with the
one-unseen and both-unseen settings, this setting is less challenging because
drug-level information is already observed during training. As a result, the
baseline models already achieve strong performance, leaving limited room for
additional improvement.

AIM-DDI still improves ACC, F1, Precision, and Recall across all three base DDI
prediction models. However, the magnitude of improvement is smaller than in the
unseen-drug settings, and AUC/AUPR slightly decrease for GIL-DDI and MKG-FENN.
This supports our main observation that AIM-DDI is most beneficial when
generalization to unseen drugs is required.

\begin{table}[tb]
\centering
\scriptsize
\setlength{\tabcolsep}{2.2pt}
\caption{
Performance comparison on the seen-drug setting.
Red and blue denote improved and decreased performance relative to the baseline, respectively.
}
\label{tab:seen_drug_results}
\resizebox{\linewidth}{!}{%
\begin{tabular}{llcccccc}
\toprule
Model & Method & ACC & AUC & AUPR & F1 & PRE & REC \\
\midrule

\multirow{3}{*}{DDIMDL}
& Baseline   
& 0.8800 $\pm$ 0.0050
& 0.9976 $\pm$ 0.0002
& 0.9318 $\pm$ 0.0033
& 0.7581 $\pm$ 0.0200
& 0.8346 $\pm$ 0.0115
& 0.7150 $\pm$ 0.0273 \\
& + AIM-DDI
& 0.9068 $\pm$ 0.0319
& 0.9987 $\pm$ 0.0011
& 0.9612 $\pm$ 0.0290
& 0.8600 $\pm$ 0.0727
& 0.8806 $\pm$ 0.0730
& 0.8513 $\pm$ 0.0676 \\
\cmidrule(lr){2-8}
& Change (\%) 
& \pos{+3.05\%} 
& \pos{+0.11\%} 
& \pos{+3.16\%} 
& \pos{+13.44\%} 
& \pos{+5.51\%} 
& \pos{+19.06\%} \\

\midrule

\multirow{3}{*}{GIL-DDI}
& Baseline
& 0.9500 $\pm$ 0.0037
& 0.9991 $\pm$ 0.0002
& 0.9830 $\pm$ 0.0016
& 0.9207 $\pm$ 0.0166
& 0.9361 $\pm$ 0.0190
& 0.9107 $\pm$ 0.0175 \\
& + AIM-DDI
& 0.9536 $\pm$ 0.0016
& 0.9959 $\pm$ 0.0007
& 0.9720 $\pm$ 0.0030
& 0.9303 $\pm$ 0.0158
& 0.9508 $\pm$ 0.0151
& 0.9181 $\pm$ 0.0185 \\
\cmidrule(lr){2-8}
& Change (\%) 
& \pos{+0.38\%} 
& \negc{-0.32\%} 
& \negc{-1.12\%} 
& \pos{+1.04\%} 
& \pos{+1.57\%} 
& \pos{+0.81\%} \\

\midrule

\multirow{3}{*}{MKG-FENN}
& Baseline   
& 0.9477 $\pm$ 0.0021
& 0.9992 $\pm$ 0.0003
& 0.9825 $\pm$ 0.0021
& 0.9158 $\pm$ 0.0192
& 0.9326 $\pm$ 0.0182
& 0.9050 $\pm$ 0.0191 \\
& + AIM-DDI
& 0.9569 $\pm$ 0.0035
& 0.9965 $\pm$ 0.0008
& 0.9768 $\pm$ 0.0032
& 0.9301 $\pm$ 0.0096
& 0.9461 $\pm$ 0.0127
& 0.9209 $\pm$ 0.0148 \\
\cmidrule(lr){2-8}
& Change (\%) 
& \pos{+0.97\%} 
& \negc{-0.27\%} 
& \negc{-0.58\%} 
& \pos{+1.56\%} 
& \pos{+1.45\%} 
& \pos{+1.76\%} \\

\bottomrule
\end{tabular}
}
\end{table}

\section{Ablation Study: Variability of LLM-Derived Semantic Modality Analysis}
\label{app:std_ablation_modality}
To complement the semantic modality ablation in Section~\ref{sec:ablation_LLM_modal}, we report the mean and standard deviation across runs in Table~\ref{tab:std_llm_ablation}.
This analysis provides an estimate of the variability of different LLM-derived semantic modality configurations in the both-unseen setting.

Overall, BioRel + MolSub remains the strongest configuration on average, achieving the best mean performance across all reported metrics.
However, several metrics show non-negligible variance across runs, especially for AUPR and macro-F1.
This suggests that while selected semantic modalities consistently provide useful complementary information, performance can still vary under the challenging both-unseen setting.
F-rank is computed from the mean performance across metrics and is therefore reported without standard deviation.

\begin{table}[tb]
\centering
\small
\caption{
Mean and standard deviation of semantic modality ablation performance.
Results are reported as mean $\pm$ standard deviation.
}
\label{tab:std_llm_ablation}
\resizebox{\linewidth}{!}{
\begin{tabular}{lccc|ccccccc}
\toprule
Variant & BioRel & MolSub & DDIGraph & ACC & AUC & AUPR & F1 & Pre & Rec & F-rank \\
\midrule
None 
&  &  &  
& 0.5046 $\pm$ 0.0714 
& 0.9492 $\pm$ 0.0178 
& 0.4560 $\pm$ 0.0860 
& 0.2397 $\pm$ 0.0657 
& 0.3001 $\pm$ 0.0713 
& 0.2272 $\pm$ 0.0732 
& 6.50 \\

BioRel 
& $\checkmark$ &  &  
& 0.4613 $\pm$ 0.0855 
& 0.9411 $\pm$ 0.0130 
& 0.4096 $\pm$ 0.0858 
& 0.2185 $\pm$ 0.0633 
& 0.2557 $\pm$ 0.0697 
& 0.2198 $\pm$ 0.0705 
& 4.67 \\

MolSub 
&  & $\checkmark$ &  
& 0.5309 $\pm$ 0.0812 
& 0.9560 $\pm$ 0.0139 
& 0.4917 $\pm$ 0.1009 
& 0.2740 $\pm$ 0.0645 
& 0.3666 $\pm$ 0.0763 
& 0.2472 $\pm$ 0.0652 
& 8.00 \\

DDIGraph 
&  &  & $\checkmark$ 
& 0.3637 $\pm$ 0.0445 
& 0.9228 $\pm$ 0.0266 
& 0.2927 $\pm$ 0.0544 
& 0.1516 $\pm$ 0.0305 
& 0.1719 $\pm$ 0.0337 
& 0.1566 $\pm$ 0.0325 
& 2.00 \\

BioRel + DDIGraph 
& $\checkmark$ &  & $\checkmark$ 
& 0.3263 $\pm$ 0.0463
& 0.9127 $\pm$ 0.0176
& 0.2607 $\pm$ 0.0426
& 0.1536 $\pm$ 0.0468
& 0.1911 $\pm$ 0.0511
& 0.1534 $\pm$ 0.0560
& 1.33 \\

MolSub + DDIGraph 
&  & $\checkmark$ & $\checkmark$ 
& 0.3640 $\pm$ 0.0546
& 0.9219 $\pm$ 0.0146
& 0.2841 $\pm$ 0.0867
& 0.1680 $\pm$ 0.0438
& 0.2086 $\pm$ 0.0507
& 0.1638 $\pm$ 0.0448
& 2.67 \\

All semantic 
& $\checkmark$ & $\checkmark$ & $\checkmark$ 
& 0.4591 $\pm$ 0.0773
& 0.9472 $\pm$ 0.0161
& 0.4100 $\pm$ 0.0809
& 0.1972 $\pm$ 0.0660
& 0.2512 $\pm$ 0.0812
& 0.1873 $\pm$ 0.0628
& 4.33 \\

Parallel GNN+LLM 
& $\checkmark$ & $\checkmark$ &  
& 0.4688 $\pm$ 0.0775
& 0.9474 $\pm$ 0.0164
& 0.4277 $\pm$ 0.0807
& 0.2468 $\pm$ 0.0788
& 0.3140 $\pm$ 0.0831
& 0.2369 $\pm$ 0.0797
& 6.50 \\

\textbf{BioRel + MolSub$^{*}$} 
& $\checkmark$ & $\checkmark$ &  
& \textbf{0.5702} $\pm$ 0.0814
& \textbf{0.9606} $\pm$ 0.0077
& \textbf{0.5085} $\pm$ 0.0962
& \textbf{0.3267} $\pm$ 0.0693
& \textbf{0.4032} $\pm$ 0.0724
& \textbf{0.3015} $\pm$ 0.0729
& \textbf{9.00} \\
\bottomrule
\end{tabular}
}
\vspace{1mm}
\footnotesize{
$^{*}$BioRel + MolSub is used as the proposed AIM-DDI configuration.
F-rank denotes the average rank across metrics and does not include standard deviation.
}
\end{table}

\section{Ablation Study: Variability of Semantic Encoder Analysis}
\label{app:std_ablation_LLM}
We further examine whether the semantic modality results depend strongly on the choice of biomedical language encoder.
Table~\ref{tab:std_semantic_encoder} compares Meditron-7B and BioBERT under the same both-unseen setting and semantic modality configuration.

The two encoders show comparable performance.
Meditron-7B achieves slightly higher mean ACC and AUC, while BioBERT achieves slightly higher mean F1.
Given the overlapping standard deviations, these results suggest that AIM-DDI is not highly sensitive to the specific biomedical encoder used for semantic representation extraction in this setting.
This supports the broader role of LLM-derived semantic modalities as complementary signals rather than a result tied to a single encoder.




\begin{table}[tb]
\centering
\small
\caption{
Mean and standard deviation of semantic encoder analysis performance.
Results are reported as mean $\pm$ standard deviation.
}
\label{tab:std_semantic_encoder}
\begin{tabular}{lccc}
\toprule
Encoder & ACC & AUC & F1 \\
\midrule
Meditron-7B & 0.5702 $\pm$ 0.0814 & 0.9606 $\pm$ 0.0077 & 0.3267 $\pm$ 0.0693 \\
BioBERT & 0.5686 $\pm$ 0.0808 & 0.9582 $\pm$ 0.0079 & 0.3296 $\pm$ 0.0664 \\
\bottomrule
\end{tabular}
\end{table}

\section{Details of Additional DrugBank-based Frameworks}
\label{app:additional_frameworks}

\paragraph{KnowDDI-based setting.}
KnowDDI~\citep{wang2024knowddi} is a knowledge-aware DDI prediction framework
that uses public DDI benchmark datasets together with an external biomedical
knowledge graph. In its 
DrugBank setting, drug-drug relations are formulated as
multi-class pharmacological interaction types. Following the KnowDDI protocol,
each drug pair is filtered to have a single relation label, and the resulting
DDI triplets are split into training, validation, and test sets.

KnowDDI further uses Hetionet~\citep{himmelstein2017hetionet} as an external biomedical knowledge graph to
provide structured drug-related information.
To reduce information leakage, drug-drug edges from the validation and test
sets are excluded from the external graph used during training. This setting
therefore evaluates whether AIM-DDI can be adapted to a knowledge-aware
DrugBank-based framework that combines DDI labels with external biomedical
knowledge.

\paragraph{KGDB-DDI-based setting.}
KGDB-DDI~\citep{zhao2025kgdbddi} constructs DrugBank-based DDI data from DrugBank v5.1.12 and KEGG~\citep{kanehisa2000kegg}, then combines
DDI pairs with biological knowledge graph information and drug background data.
The biological knowledge graph includes drug-related enzymes, categories,
pathways, targets, and gene information. In addition, KGDB-DDI uses drug
background descriptions, such as research and development background,
indications, efficacy, and mechanism of action.

In the DrugBank setting, known DDIs are used as positive samples, and the same
number of negative samples are randomly selected to construct the final dataset.
The data are split into training, validation, and test sets with an 8:1:1 ratio.
The biological knowledge graph used in this setting does not contain explicit
DDI edges, which helps reduce information leakage. This setting evaluates
whether AIM-DDI can be adapted to a DrugBank-based framework that combines
knowledge graph features with drug background information.

\section{Implementation Details}
\label{app:implementation_details}

All experiments are implemented in PyTorch and conducted on a machine equipped
with a single NVIDIA GeForce RTX 5090 GPU with 32GB memory. We use 10-fold
cross-validation for the main DrugBank-based benchmark. For the additional
KnowDDI- and KGDB-DDI-based settings, we follow the evaluation protocols of the
corresponding frameworks. For reproducibility, we fix Python, NumPy, and PyTorch
random seeds and use deterministic CUDA settings. For one full 10-fold run of
the main DrugBank-based experiments, the measured runtimes were 1 h 59 min 19 s
for DDIMDL, 1 h 18 min 29 s for GIL-DDI, and 24 min 21 s for MKG-FENN.

Table~\ref{tab:hyperparameters} shows selected hyperparameters of each backbone model.
For all AIM-DDI variants, the adaptive fusion module uses four experts, top-2
routing, expert-choice routing, four attention heads, attention dropout of 0.1,
and a capacity factor of 1.25. Unless otherwise specified, focal loss is used
with $\gamma=2.0$ for imbalanced DDI event classification.

\begin{table}[tb]
\centering
\small
\caption{Selected hyperparameters for each backbone model.}
\label{tab:hyperparameters}
\begin{tabular}{lccc}
\toprule
Configuration & DDIMDL & GIL-DDI & MKG-FENN \\
\midrule
Epochs & 120 & 120 & 120 \\
Batch size & 128 & 1024 & 1024 \\
Learning rate & $1\times10^{-3}$ & $1\times10^{-3}$ & $5\times10^{-3}$ \\
Weight decay & -- & $1\times10^{-8}$ & $1\times10^{-8}$ \\
Dropout & 0.3 & 0.3 & 0.5 \\
Hidden dimension & 256 & 256 & 256 \\
Neighbor sample size & -- & 6 & 6 \\
Number of experts & 4 & 4 & 4 \\
Top-$k$ routing & 2 & 2 & 2 \\
Attention heads & 4 & 4 & 4 \\
Attention dropout & 0.1 & 0.1 & 0.1 \\
Capacity factor & 1.25 & 1.25 & 1.25 \\
Routing type & expert-choice & expert-choice & expert-choice \\
Focal loss $\gamma$ & 2.0 & 2.0 & 2.0 \\
\bottomrule
\end{tabular}
\end{table}

DDIMDL uses feature-based pair representations constructed from SMILES,
target, and enzyme information. Each pair-level feature is encoded by a
lightweight encoder and then passed to the AIM-DDI fusion module. GIL-DDI and
MKG-FENN use graph-based drug representations with a neighbor sample size of 6.

For evaluation under unseen-drug settings, drug representations for test pairs
are constructed using available graph structures and semantic information.
When test-time adjacency is enabled, unseen-drug representations are obtained
from the most similar training drugs based on the corresponding semantic
embeddings.

For semantic modalities, we use precomputed per-drug embeddings for each
modality. DDIMDL and GIL-DDI support up to four semantic sources, whereas
MKG-FENN uses three semantic sources. In the proposed configuration, BioRel
and MolSub semantic sources are used unless otherwise specified.




\section{Ablation Study: Expert Routing and Modality Contribution Details}
\label{app:expert_routing_details}

We analyze the behavior of the expert-choice fusion module in the
both-unseen setting using the MKG-FENN + AIM-DDI configuration.
For each drug-pair instance, we record the assigned expert, gate
probability, DDI event label, and modality contribution scores. This analysis
examines whether the adaptive expert fusion module uses multiple experts and
whether different experts show different event and modality patterns.

\paragraph{Expert utilization.}
Table~\ref{tab:expert_sample_counts} reports the number of samples assigned to
each expert. Samples are distributed across all four experts, indicating that
the routing mechanism does not collapse to a single expert.

\begin{table}[b]
\centering
\small
\caption{
Number of samples assigned to each expert in the both-unseen setting
using MKG-FENN + AIM-DDI.
}
\label{tab:expert_sample_counts}
\begin{tabular}{lc}
\toprule
Expert & Assigned samples \\
\midrule
Expert 0 & 802 (21.56\%)\\
Expert 1 & 1,059 (28.47\%)\\
Expert 2 & 1,107 (29.76\%)\\
Expert 3 & 752 (20.22\%)\\
\bottomrule
\end{tabular}
\end{table}

\paragraph{Gate confidence.}
Table~\ref{tab:expert_gate_probability} summarizes the mean gate probabilities
grouped by assigned expert. The diagonal entries are close to one, showing that
samples assigned to each expert receive the highest mean gate probability for
that expert. This indicates confident and expert-specific routing behavior.

\begin{table}[tb]
\centering
\small
\caption{
Mean gate probability grouped by assigned expert. Rows indicate assigned
experts and columns indicate gate probability dimensions.
}
\label{tab:expert_gate_probability}
\begin{tabular}{lcccc}
\toprule
Assigned expert & Gate 0 & Gate 1 & Gate 2 & Gate 3 \\
\midrule
Expert 0 & \textbf{0.9892} & 0.0048 & 0.0041 & 0.0019 \\
Expert 1 & 0.0033 & \textbf{0.9919} & 0.0029 & 0.0019 \\
Expert 2 & 0.0036 & 0.0041 & \textbf{0.9905} & 0.0017 \\
Expert 3 & 0.0014 & 0.0032 & 0.0029 & \textbf{0.9925} \\
\bottomrule
\end{tabular}
\end{table}

\paragraph{Expert-wise DDI event distribution.}
Table~\ref{tab:expert_top_events} reports the top DDI event classes assigned to
each expert. Frequent event classes appear across multiple experts, reflecting
the imbalanced nature of the 65-class DDI prediction task. However, the relative
concentration of event classes differs across experts, suggesting partial
specialization over event types.

\begin{table}[tb]
\centering
\small
\caption{
Top DDI event classes assigned to each expert in the both-unseen setting.
The ratio denotes the proportion of samples assigned to the corresponding event
class within each expert.
}
\label{tab:expert_top_events}
\begin{tabular}{lccc}
\toprule
Expert & Top-1 event & Top-2 event & Top-3 event \\
\midrule
Expert 0 & Class 0 (294, 36.66\%) & Class 1 (171, 21.32\%) & Class 2 (162, 20.20\%) \\
Expert 1 & Class 0 (307, 28.99\%) & Class 1 (189, 17.85\%) & Class 2 (146, 13.79\%) \\
Expert 2 & Class 1 (259, 23.40\%) & Class 0 (257, 23.22\%) & Class 2 (209, 18.88\%) \\
Expert 3 & Class 1 (307, 40.82\%) & Class 0 (124, 16.49\%) & Class 5 (60, 7.98\%) \\
\bottomrule
\end{tabular}
\end{table}

\begin{table}[t!]
\centering
\small
\caption{
Expert-wise modality contribution in the both-unseen setting. BioRel denotes
LLM-derived drug-protein biological relation context, MolSub denotes
LLM-derived molecular substructure context, and DDIGraph denotes the
graph-based drug-drug interaction representation.
}
\label{tab:expert_modality_contribution}
\begin{tabular}{lccc}
\toprule
Expert &  BioRel & MolSub & DDIGraph \\
\midrule
Expert 0 & 0.1425 & 0.0555 & 0.8021 \\
Expert 1 &  0.1276 & 0.0699 & 0.8024 \\
Expert 2 & 0.1199 & 0.0730 & 0.8071 \\
Expert 3 & 0.1126 & 0.0822 & 0.8051 \\
\bottomrule
\end{tabular}
\end{table}

\begin{table}[t!]
\centering
\small
\caption{
Detailed token-level modality contribution for each expert. Each value denotes
the average contribution score of the corresponding modality token among samples
assigned to the expert.
}
\label{tab:expert_token_contribution}
\resizebox{\linewidth}{!}{
\begin{tabular}{lccccccc}
\toprule
Expert & drugA-BioRel & drugA-MolSub & drugA-DDIGraph & drugB-BioRel & drugB-MolSub & drugB-DDIGraph \\
\midrule
Expert 0 &  0.0732 & 0.0235 & 0.3883 & 0.0693 & 0.0320 & 0.4138 \\
Expert 1 & 0.0670 & 0.0331 & 0.3870 & 0.0607 & 0.0369 & 0.4155 \\
Expert 2 &  0.0645 & 0.0292 & 0.3951 & 0.0554 & 0.0439 & 0.4120 \\
Expert 3 & 0.0586 & 0.0278 & 0.4123 & 0.0541 & 0.0544 & 0.3928 \\
\bottomrule
\end{tabular}
}
\end{table}

\paragraph{Expert-wise modality contribution.}

We analyze how each expert uses the available modality tokens.
For each expert, we aggregate the contribution scores of BioRel, MolSub, and DDIGraph over the two drugs in a pair.
As shown in Table~\ref{tab:expert_modality_contribution}, DDIGraph receives the largest contribution across all experts, indicating that the graph-based DDI representation remains the primary structural signal.
This is consistent with the role of DDI topology in the MKG-FENN baseline.

Importantly, AIM-DDI does not discard this strong baseline signal, but augments it with complementary semantic information.
BioRel and MolSub receive smaller but non-negligible contributions, and their relative weights vary across experts.
For example, BioRel is highest in Expert 0, whereas MolSub is highest in Expert 3.
This suggests that the expert fusion module preserves the dominant DDI-graph signal while allowing different experts to incorporate semantic modalities to different degrees.

\paragraph{Token-level modality contribution.}
Table~\ref{tab:expert_token_contribution} further breaks down the contribution scores by drug side and modality token.
The DDIGraph tokens of both drugs consistently receive the largest scores, showing that interaction-topology information is used symmetrically from the two drugs in a pair.
At the same time, the semantic tokens show expert-specific patterns. BioRel contributions gradually decrease from Expert 0 to Expert 3, while MolSub contributions increase, especially for the drugB-MolSub token.
These patterns suggest that experts specialize not by replacing the DDI-graph signal, but by emphasizing different complementary semantic contexts around it.

\section{Ablation Study: Sensitivity Analysis on the Number of Experts}
\label{app:expert_number}

We analyze the effect of the number of experts in the AIM-DDI fusion module
under the both-unseen setting. In this experiment, only the number of experts is
varied, while the remaining configuration is kept fixed. The purpose of this
analysis is to examine how sensitive AIM-DDI is to the capacity of the adaptive
expert fusion module.

As shown in Table~\ref{tab:expert_number}, performance does not improve monotonically as the number of experts increases.
The two-expert configuration achieves the highest ACC, F1, Precision, and Recall, while the four-expert configuration used in AIM-DDI achieves the highest AUC and remains close to the best configuration on the other metrics.
Overall, the performance differences among 2-5 experts are relatively small, suggesting that AIM-DDI is not highly sensitive to the exact number of experts within this range.
We use four experts in the main experiments as a moderate-capacity setting that provides competitive performance across metrics.

\begin{table}[tb]
\centering
\scriptsize
\setlength{\tabcolsep}{2.5pt}
\caption{
Sensitivity analysis on the number of experts in the both-unseen setting.
Results are reported as mean $\pm$ standard deviation.
}
\label{tab:expert_number}
\resizebox{\linewidth}{!}{%
\begin{tabular}{lcccccc}
\toprule
Experts & ACC & AUC & AUPR & F1 & PRE & REC \\
\midrule
2 
& \textbf{0.5715} $\pm$ 0.0684
& 0.9566 $\pm$ 0.0096
& 0.5106 $\pm$ 0.0742
& \textbf{0.3314} $\pm$ 0.0735
& \textbf{0.4039} $\pm$ 0.0679
& \textbf{0.3061} $\pm$ 0.0775 \\

3 
& 0.5589 $\pm$ 0.0686
& 0.9598 $\pm$ 0.0073
& \textbf{0.5121} $\pm$ 0.0923
& 0.3191 $\pm$ 0.0527
& 0.3894 $\pm$ 0.0525
& 0.3013 $\pm$ 0.0548 \\

4 (Ours) 
& 0.5702 $\pm$ 0.0814
& \textbf{0.9606} $\pm$ 0.0077
& 0.5085 $\pm$ 0.0962
& 0.3267 $\pm$ 0.0693
& 0.4032 $\pm$ 0.0724
& 0.3015 $\pm$ 0.0729 \\

5 
& 0.5610 $\pm$ 0.0795
& 0.9604 $\pm$ 0.0053
& 0.5091 $\pm$ 0.0927
& 0.3178 $\pm$ 0.0851
& 0.3889 $\pm$ 0.0920
& 0.2957 $\pm$ 0.0830 \\
\bottomrule
\end{tabular}
}
\end{table}

\section{Ablation Study: Pair Representation Construction}
\label{app:pair_representation}



We further analyze how drug-pair representations affect modality-token integration.
This ablation is conducted in the both-unseen setting using the MKG-FENN-based AIM-DDI implementation.
All variants use the same semantic modality configuration, training objective, and inference setting.

We compare three pair representation strategies.
First, the current AIM-DDI formulation keeps drug-specific modality tokens separate:
\begin{equation}
    \mathbf{H}^{\mathrm{sep}}_{uv}
    =
    [
    \mathbf{h}^{(1)}_u;
    \mathbf{h}^{(2)}_u;
    \mathbf{h}^{(3)}_u;
    \mathbf{h}^{(1)}_v;
    \mathbf{h}^{(2)}_v;
    \mathbf{h}^{(3)}_v
    ].
\end{equation}
Second, the drug-average variant first averages modality representations within each drug and then concatenates the two drug-level vectors:
\begin{equation}
    \mathbf{H}^{\mathrm{avg}}_{uv}
    =
    \left[
    \frac{1}{M}\sum_{m=1}^{M}\mathbf{h}^{(m)}_u
    \;\Big\|\;
    \frac{1}{M}\sum_{m=1}^{M}\mathbf{h}^{(m)}_v
    \right].
\end{equation}
Third, the modality-pair variant constructs one token per modality by concatenating the two drug representations within the same modality and projecting them through an MLP:
\begin{equation}
    \mathbf{q}^{(m)}_{uv}
    =
    \mathrm{MLP}^{(m)}_{\mathrm{pair}}
    \left(
    \left[
    \mathbf{h}^{(m)}_u
    \;\Big\|\;
    \mathbf{h}^{(m)}_v
    \right]
    \right),
    \qquad
    \mathbf{H}^{\mathrm{pair}}_{uv}
    =
    [
    \mathbf{q}^{(1)}_{uv};
    \cdots;
    \mathbf{q}^{(M)}_{uv}
    ].
\end{equation}

\begin{table}[tb]
\centering
\scriptsize
\setlength{\tabcolsep}{2.5pt}
\caption{
Effect of drug-pair representation construction in the both-unseen setting.
All variants use the same semantic modality configuration and differ only
in how drug-level modality representations are converted into pair representations.
Results are reported as mean $\pm$ standard deviation.
}
\label{tab:pair_representation}
\resizebox{\linewidth}{!}{%
\begin{tabular}{lcccccc}
\toprule
Pair representation & ACC & AUC & AUPR & F1 & PRE & REC \\
\midrule

\makecell[l]{\textbf{Separate drug-modality}\\ \textbf{tokens (AIM-DDI)}}
& \textbf{0.5702} $\pm$ 0.0814
& \textbf{0.9606} $\pm$ 0.0077
& 0.5085 $\pm$ 0.0962
& \textbf{0.3267} $\pm$ 0.0693
& \textbf{0.4032} $\pm$ 0.0724
& \textbf{0.3015} $\pm$ 0.0729 \\

\makecell[l]{Drug-average\\ concatenation}
& 0.5613 $\pm$ 0.0824
& 0.9526 $\pm$ 0.0047
& \textbf{0.5095} $\pm$ 0.0875
& 0.3193 $\pm$ 0.0759
& 0.3915 $\pm$ 0.0769
& 0.2951 $\pm$ 0.0762 \\

Modality-pair tokens
& 0.5487 $\pm$ 0.0740
& 0.9557 $\pm$ 0.0079
& 0.4899 $\pm$ 0.0786
& 0.2740 $\pm$ 0.0410
& 0.3236 $\pm$ 0.0465
& 0.2632 $\pm$ 0.0437 \\
\bottomrule
\end{tabular}
}
\end{table}


As shown in Table~\ref{tab:pair_representation}, the separate drug-modality
token representation used in AIM-DDI achieves the best performance on five out
of six metrics, including ACC, AUC, F1, Precision, and Recall. Although
drug-level average concatenation obtains a slightly higher AUPR, it performs
worse on the remaining metrics. The modality-wise pair-token strategy shows the
lowest overall performance. These results suggest that preserving drug-specific
modality tokens is useful for modeling cross-drug and cross-modality
interactions in the both-unseen setting.

\section{Limitations}
\label{app:limitations}

AIM-DDI is designed as a model-agnostic multimodal integration module, but its
performance still depends on the quality and availability of the input
representations produced by each backbone model. If a backbone provides weak or
noisy modality representations, the integration module may not fully compensate
for such limitations.

Second, the effectiveness of LLM-derived semantic modalities depends on the
alignment between the semantic source and the DDI prediction objective. As shown
in the semantic modality ablation, adding all available semantic sources does
not necessarily improve performance, and poorly aligned sources can degrade
prediction quality. This suggests that modality selection remains an important
factor.

Third, AIM-DDI is evaluated mainly on DrugBank-derived DDI benchmarks. Although
we additionally evaluate the method under KnowDDI- and KGDB-DDI-based settings,
these settings are still related to DrugBank-derived interaction data. Therefore,
further evaluation on independent pharmacovigilance or clinical datasets is
needed to assess broader generalization.


\section{Broader Impacts}
\label{app:broader_impacts}

This work aims to improve computational drug-drug interaction prediction,
especially for unseen drugs. Accurate prediction of potential DDIs may support
safer drug development, pharmacovigilance, and prioritization of candidate
interactions for further expert review.

At the same time, incorrect predictions may lead to inappropriate confidence in
a predicted interaction or failure to identify a harmful interaction. Therefore,
the proposed method should be used only as a decision-support or research tool,
not as a replacement for clinical judgment, regulatory evaluation, or
experimental validation.

The study uses existing biomedical benchmark datasets and does not involve
human subjects, private patient records, or personally identifiable information.
The main risk is potential misuse or overinterpretation of model predictions in
clinical or pharmaceutical settings. Clear communication of uncertainty,
validation requirements, and intended use is necessary before any real-world
deployment.


\end{document}